\documentclass[preprint,12pt,3p]{elsarticle}

\usepackage{graphics}
\usepackage{graphicx}
\usepackage{epsfig}

\usepackage{amssymb}
\usepackage{xcolor}
\usepackage{times}
\usepackage{epsfig}
\usepackage{graphicx}
\usepackage{subfig}
\usepackage{float}
\usepackage{amsmath}
\usepackage{amssymb}
\usepackage{algorithm}
\usepackage{algorithmic}
\usepackage{multirow}
\usepackage{setspace}

\makeatletter \@fpsep = 2pt \makeatother
\begin{document}

\begin{spacing}{1.0}

\begin{frontmatter}

\title{Image super-resolution with an enhanced group convolutional neural network}

\author[label1,label2,label3]{Chunwei Tian}
\author[label3]{Yixuan Yuan\corref{cor1}}
\ead{yxyuan.ee@cityu.edu.hk}
\author[label4]{Shichao Zhang}
\author[label5]{Chia-Wen Lin\corref{cor1}}
\ead{cwlin@ee.nthu}
\cortext[cor1]{Corresponding author}
\author[label6,label7]{Wangmeng Zuo}
\author[label8,label9]{David Zhang}

\address[label1]{School of Software, Northwestern Polytechnical University, Xi’an, Shaanxi, 710129, China}
\address[label2]{National Engineering Laboratory for Integrated Aero-Space-Ground-Ocean Big Data Application Technology, Xi’an, Shaanxi, 710129, China}
\address[label3]{Department of Electrical Engineering, City University of Hong Kong, Hong Kong SAR, China}
\address[label4]{School  of  Computer  Science  and  Engineering,  Central  South  University,  Changsha,  Hunan, 410083, China}
\address[label5]{Department of Electrical Engineering and the Institute of Communications Engineering, National Tsing Hua University, Hsinchu, Taiwan.}
\address[label6]{School of Computer Science and Technology, Harbin Institute of Technology, Harbin, Heilongjiang, 150001, China}
\address[label7]{Peng Cheng Laboratory, Shenzhen, Guangdong, 518055, China.}
\address[label8]{School of Data Science, The Chinese University of Hong Kong (Shenzhen), Shenzhen, Guangdong, 518172, China}
\address[label9]{Shenzhen Institute of Artificial Intelligence and Robotics for Society, Shenzhen, China}

\begin{abstract}
CNNs with strong learning abilities are widely chosen to resolve super-resolution problem. However, CNNs depend on deeper network architectures to improve performance of image super-resolution, which may increase computational cost in general. In this paper, we present an enhanced super-resolution group CNN (ESRGCNN) with a shallow architecture by fully fusing deep and wide channel features to extract more accurate low-frequency information in terms of correlations of different channels in single image super-resolution (SISR). Also, a signal enhancement operation in the ESRGCNN is useful to inherit more long-distance contextual information for resolving long-term dependency. An adaptive up-sampling operation is gathered into a CNN to obtain an image super-resolution model with low-resolution images of different sizes. Extensive experiments report that our ESRGCNN surpasses the state-of-the-arts in terms of SISR performance, complexity, execution speed, image quality evaluation and visual effect in SISR. Code is found at https://github.com/hellloxiaotian/ESRGCNN.
\end{abstract}


\end{frontmatter}


\section{Introduction}
\label{sec-1}

Image super-resolution (SR) technique devotes to  recover a clearer image from an unclear observation through a classical equation $y = {x_{ \downarrow s}}$,  where $x$ is a high-definition (also treated high-resolution, HR) image, $y$ denotes a unclear (also regraded as low-resolution, LR) image and $s$ denotes a given scale factor. Specifically, the same LR image can be obtained from numerous HR images by a down-sampling operation, according to the mentioned equation. That is, SISR problem does not have unique solution, which can be known as an ill-posed inverse problem \cite{hui2018fast,tian2020coarse}. To address this problem, scholars present a lot of single image SR (SISR) methods \cite{deng2015single}. For instance, Hong et al. divided LR-HR pairs of patches into different clusters, then made a fuzzy rule in image super-resolution, according to these clusters\cite{purkait2014fuzzy}. Liu et al. improved a weighted random forest model with rotation in image super-resolution \cite{liu2017image}. There are some effective in image super-resolution methods, i.e., interpolation-based techniques \cite{ismail2021sparse}, sparse based dictionary learning techniques  \cite{zhang2012image}, neighbor embedding techniques  \cite{chang2004super} and Bayesian technique \cite{zhang2012generative}. Although these methods have obtained excellent performance of image super-resolution,  
some of these methods may drop detailed information, which limited effect in SISR performance \cite{wang2020deep,yang2019deep}. Also, due to manual tuning, most of these methods are not flexible. Additionally, they usually rely to complex optimization algorithms for boosting SISR performance, which would decrease efficiency of SISR.

Recently, due to plug-in network architectures and flexible training mechanisms, deep networks have strong self-learning ability to gain better restoration performance \cite{hui2021progressive,liang2020video,zhou2021cross}. For instance, Cui et al. used multiple stacked
auto-encoders with non-local self-similarity in image super-resolution \cite{cui2014deep}. Dong et al. presented a 3-layer CNN model known as SRCNN via pixel mapping strategy to recover a high-quality image \cite{dong2015image}. Although the SRCNN obtained more effective super-resolution (SR) results in comparison with traditional SR methods, it had slow convergence speed and large training cost. To overcome this problem, deeper network architectures are designed to pursue excellent SR effect. For instance, Kim et al. designed a deeper network architecture named VDSR through stacking a series of convolutional layers, residual learning (RL) techniques and gradient clipping operation to accelerate training of SR model \cite{kim2016accurate}.  Since then, enhancing the effect of local information from different layers by multiple use of RL techniques becomes popular to further ease training difficulty and promote the SR performance. A deep recursive residual network (DRRN) fused RL and recursive learning strategies to improve the generalization ability of a SR method \cite{tai2017image}. Specifically, this RL technique can mitigate training difficulty of SR model and the recursive learning technique can make a trade-off between network depth and parameters. Alternatively, a residual encoder-decoder network (RED) connected convolutional and deconvolutional layers through utilizing RL techniques to construct symmetrical CNN for predicting HR images \cite{mao2016image}. Besides, a deeper memory network (MemNet) used RL technique to mine different level features for enhancing the influence of prior layer in SISR \cite{tai2017memnet}. Using signal processing idea (i.e., wavelet transform) into CNN can achieve prominent performance for SISR \cite{liu2018multi}.
The combination of a deep CNN and wavelet idea can obtain more detailed information to improve quality of predicted images \cite{guo2017deep}. Although these methods can recover high-definition images, they had a high computational costs by using bicubic operation for constructing input of SR network \cite{tian2020lightweight}. To overcome the mentioned drawback, scholars directly used LR images as input of SR network to predict a relation from given unclear images to high-definition images. As the pioneer, Dong et al. \cite{dong2016accelerating} firstly used an up-sampling operation to amplify the resolution of obtained low-frequency features, which can accelerate training speed with degrading visual effect.

Making full use of hierarchical features from different network layers can enhance robustness of low-frequency features to enhance the resolution of predicting SR images. For instance, Lim et al. presented an enhanced deep network known enhanced deep SR network (EDSR) through utilizing multi-scale techniques to fuse different low-frequency features for improving visual results \cite{lim2017enhanced}. Besides, Zhang et al. exploited different filter sizes to mine different features, then fused these features to generate more accurate features in SR \cite{zhang2018residual}.Although these methods have achieved excellent SR results, their deeper  network architectures suffered from bigger computational costs. Additionally, these SR models only dealt with a single scale by one model, which cannot satisfy requirements of real applications. In this paper, we present an enhanced super-resolution group CNN (ESRGCNN) through mainly stacking six group enhanced convolutional blocks (GEBs), connecting a combination of a convolution and an activation function, an adaptive up-sampling mechanism and a single convolutional layer. Specifically, the GEB uses group convolutions and RL techniques to enhance expressive ability of obtained low-frequency information in terms of correlations of different channels for balancing SISR performance and complexity. Also, this combination can prevent over enhancement of obtained low-frequency information. A signal enhancement operation in GEB can inherit long-distance contextual information from shallow layers via skip connection operation to offer complementary information for deep layers, which is useful to deal with long-term dependence ability. Also, an adaptive up-sampling mechanism is utilized to obtain a super-resolution feature mapping from LR frequency features to HR frequency features for different scales, which satisfies requirement of SR technique on digital devices. Additionally, the final convolutional layer is used to construct a HR image.

This main contributions of our ESRGCNN can be reported as follows.

(1)	The proposed 40-layer ESRGCNN uses group convolutions and residual operations to enhance deep and wide correlations of different channels to implement an efficient SR network. 

(2) An adaptive up-sampling mechanism is used to obtain a flexible SR model, which is very beneficial to real applications.
	
(3) Shallow ESRGCNN only uses the number of parameters to $5.6\%$ of 134-layer RDN and $9.6\%$ of 384-layer CSFM to obtain excellent visual effects, which also takes the running time to $3\%$ of popular RDN and CSFM in recovering a HR image of $1024 \times 1024$.

The remaining parts of this paper have the following introduce. The second section lists related work. The third section gives  ESRGCNN. The fourth section describes experimental results. The final section summaries the whole paper.
\section{Related work}
\subsection{Deep CNNs based group convolutions for SISR}
Numerous deep learning methods with strong self-learning ability have been used in SISR \cite{zhang2018image,tian2020deep}. Since most of these methods are treated equally across channels to enhance the effect of hierarchical features for SISR, that hinders expressive ability of CNN \cite{zhang2018residual}. To resolve this issue, deep CNNs based group convolutions are developed in SISR. That roughly includes two kinds in general, such as attention based channels and features fusion based channels.

The first category used attention techniques to strengthen the effect of key channels for boosting the SISR performance and speed. For instance, Hu et al. merged channel-wise attention and spatial attention into deeper network to extract prominent low-frequency information for promoting SISR performance \cite{hu2019channel}. Besides, Dai et al. exploited second-order feature statistics to obtain more accurate features and more discriminative representations for SISR \cite{dai2019second}.

The second category fused hierarchical channel features by using RL or concatenation operations to obtain abundant low-frequency information for SISR. For instance, Jain et al. removed useless connections via group convolutions to accelerate the speed of training SR model \cite{jain2018efficient}. Since then, to obtain more robust features, Zhao et al. relied two sub-networks channels-based to expend diversity for offering complementary features in SISR \cite{zhao2019channel}. To further reduce the complexity of the SR model, feature fusion based channel via group convolutions was presented \cite{hui2019lightweight}. For instance, Yang et al. utilized softmax feature fusion mechanism and spindle block to reduce parameters for constructing a lightweight blind SR network \cite{yang2019lightweight}. Alternatively, cascading several group convolutional networks with several convolutions can reduce complexity of SR network \cite{yang2020lightweight}. To address complex scenes, a multimodal deep learning technique uses multi-modal images of a same scene into two sub-networks to extract complementary features for improving performance of image processing task \cite{hong2020more}. Additionally, to extract sequence attributes of spectral signatures, a sequential perspective with transformers is developed to learn spectrally local sequence information. Also, a cross-layer skip connection is used to enhance memory ability of a designed network for improving performance in image classification \cite{hong2021spectralformer}. The above two methods have an important reference value for image super-resolution.

All the mentioned methods illustrate the effectiveness of performance and complexity in SISR. Therefore, we use group convolutions in this paper for SISR. Specifically, differ from other group convolutional SR methods, our method enhances correlations of different channels via two adjacent layers rather than using the current layer as inputs of later layers to extract more accurate low-frequency features and reduce complexity in SISR.

\subsection{Multilevel feature fusion for SISR}
According to previous illustrations, it is known that single feature learning methods cannot guarantee robustness of learned features for complex screens. Inspired by that, feature fusion method is developed \cite{hong2020more,song2021multi}. Due to its good performance, this method is popular for high-level computer vision 
tasks \cite{han2021automatic}. Besides, it is known that a network with a big depth may keep poorer memory ability of shallow layers. To overcome this challenge, multilevel feature fusion method in the CNNs is presented to enlarge the effect of local features for a SR task \cite{wang2020deep,lan2020cascading}. That includes two kinds in general: low-frequency feature fusion and high-frequency feature fusion.

The first category uses a bicubic method to amplify LR images for inputting the deep network and makes full use of hierarchical high-frequency features to obtain a SR model. For instance, a deeply-recursive convolutional network (DRCN) combined recursive-supervision and skip connection operations to enhance effect of obtained hierarchical information for relieving difficulty of training \cite{kim2016deeply}. Besides, the MemNet \cite{tai2017memnet} exploited recursive unit and gate unit to enhance power of the current layer for improving expressive ability of deep network in SISR. Although these methods can obtain high-quality images, they had high complexity.

The second category directly utilizes LR image as input of deep network and fully exploits hierarchical low-frequency features to obtain more accurate information with fast processing speed. Finally, this method uses up-sampling operation in the deep layer of network to magnify obtained information to construct a HR image. That is, a cascading residual network (CARN) utilized cascading block and RL operation to strength local obtained features for extracting more abundant information in SISR \cite{ahn2018fast}. Alternatively, an enhanced SR CNN (LESRCNN) fused RL technique into heterogeneous convolutional structure to reinforce low-frequency features from deep layers for achieving significant SR performance and fast speed \cite{tian2020lightweight}.

According to mentioned illustrations, it is known that multilevel feature fusion method is useful to obtain a clearer image. Because the first category method has high complexity, we use idea of the second method into this paper as Fig.1.
\section{The proposed method}
Our proposed ESRGCNN contains three convolutional layers, six group convolutional blocks known GEBs and an adaptive upsampling mechanism as shown in Fig.1. Specifically, the GEB based group convolutions and RL techniques fuse wide and deep channel features to heighten representation ability of low-frequency features for boosting SISR performance and reducing complexity, according to correlations of different channels. Also, a signal enhancement in each GEB can obtain long-distance contextual information to address long-term dependence problem of ESRGCNN. Due to shallow architecture, ESRGCNN has fast execution speed for SISR. Besides, the adaptive upsampling mechanism exploits a flexible valve to deal with LR images of different scales for achieving a flexible SR model. Thus, the proposed ESRGCNN may be better applied on real digital devices. More information of ESRGCNN is given in latter sections.
\subsection{Network architecture}
As reported in Fig.1, a 40-layer ESRGCNN mainly comprises four parts: 2-layer combination of convolutional layer and ReLU \cite{krizhevsky2012imagenet}, 36-layer GEBs, 1-layer upsampling mechanism and single convolution layer. Each combination of a convolutional layer and a ReLU is that a convolutional layer connects a ReLU, which acts head and intermediate of the whole ESRGCNN, respectively. And they are symboled as Conv+ReLU in Fig.1. Specifically, the first Conv+ReLU uses a convolutional layer to obtain key information of a given unclear image, then it connects an activation function, ReLU \cite{krizhevsky2012imagenet} to covert obtained linear features into non-linearity, where its input channel number, output channel number and filter size denote 3, 64 and $3 \times 3$, respectively. Since then, six GEBs are used to strengthen expressive ability of low-frequency features by correlations of different channels for achieving impressive results and competitive complexity of SISR, where input channel number, output channel number and filter size of each GEB are 64, 64 and $3 \times 3$, respectively. Also, a signal enhancement operation is used into each GEB to extract long-distance contextual information for tackling long-term dependence problem of deep SR network. Next, the second combination of a convolutional layer and a ReLU is used to prevent of over-enhanced of obtained feature point pixels from the mentioned operation. Specifically, its channel number, output channel number and filter size are the same as each GEB. Then, this adaptive upsampling operation uses a flexible threshold to covert extracted low-frequency information to high-frequency information for achieving a flexible SR model. Finally, the single convolutional layer as the final layer of ESRGCNN constructs a predicted SR image via obtained high-frequency features, where its channel number, output channel number and filter size are 64, 3 and $3 \times 3$, respectively. To intuitively express the mentioned process, we define some symbols as follows. We assume that given $I_{LR}$ and $I_{SR}$ denote input and output of ESRGCNN, respectively. Let $C$ and $R$ be a convolutional function and a ReLU function, respectively. $6GEB$ stands for functions of six group enhanced convolutional blocks. And $UP$ represents an upsampling operation. Therefore, the mentioned illustrations can be formulated as

\begin{small}
\begin{equation}
\begin{array}{ll}
{I_{SR}} & =  {C(UP(R(C(6GEB(R(C(I_{LR})))))))}\\
 & =   {ESRGCNN(I_{LR})},
\end{array}
\end{equation}
\end{small}
where $ESRGCNN$ stands for the function of ESRGCNN, which is optimized via loss function of Section 3.2.
\begin{figure}[!htbp]
\centering
\subfloat{\includegraphics[width=5.5in]{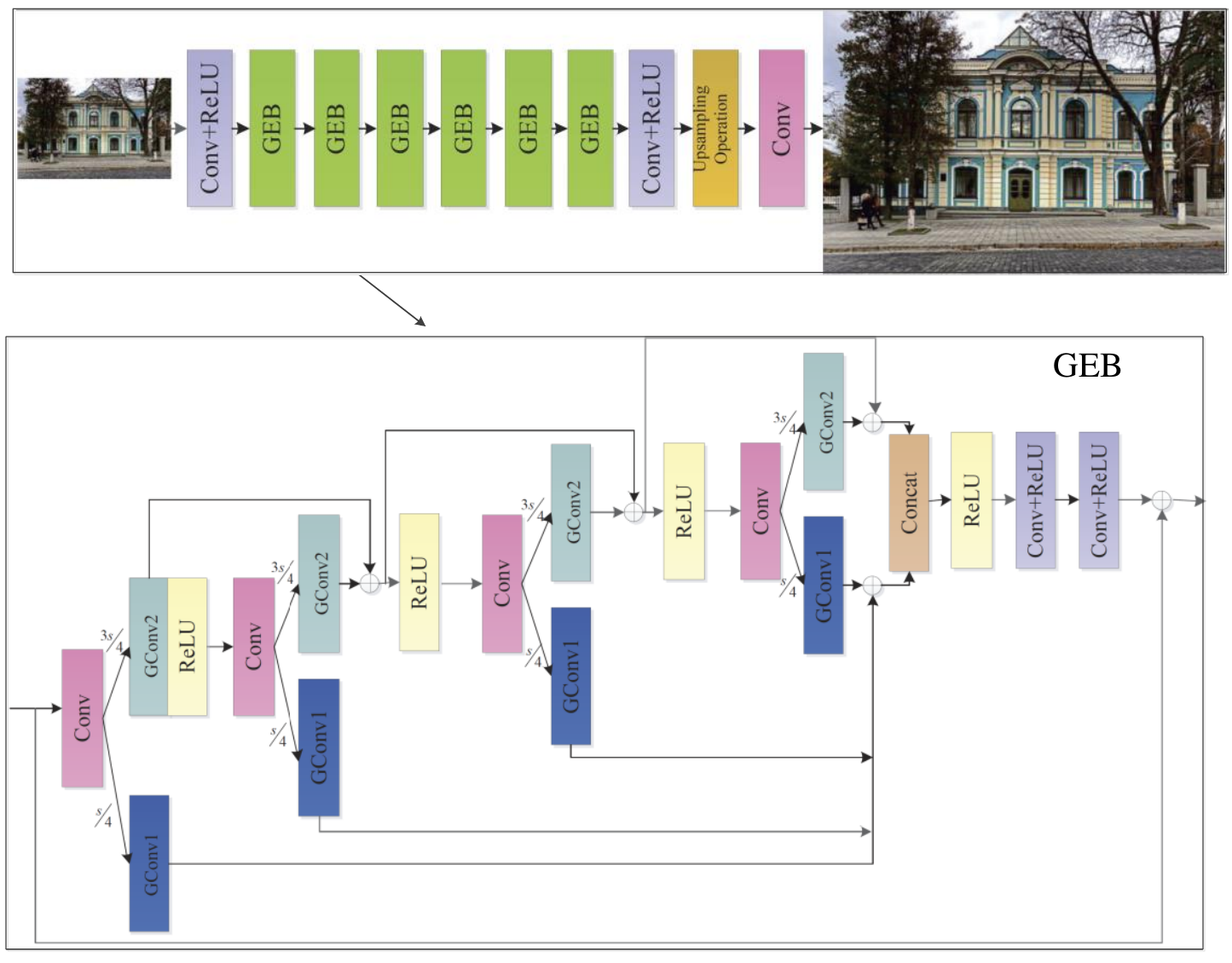}
\label{fig_second_case}}
\caption{Network architecture of ESRGCNN.}
\label{fig:7}
\end{figure}
\subsection{Loss function}
This paper chooses public mean squared error (MSE) \cite{douillard1995iterative,tian2020lightweight} as a loss function to determine optimal parameters of ESRGCNN for predicting SR images. That is, a set of pairs   $\{ I_{LR}^k,I_{HR}^k\} _{k = 1}^T$ from training dataset in Section 4.1 is used into the MSE to train a SR model, where $I_{LR}^k$ and $I_{HR}^k$ express the $kth$ low- and high-resolution image patches from training dataset, respectively. $T$ denotes the total of training image patches. The process can be shown in Eq.(2):
\begin{equation}
\begin{array}{l}
l(p) = \frac{1}{{2T}}{\sum\limits_{k = 1}^T {\left\| {{f_{ESRGCNN}}(I_{LR}^k) - I_{HR}^k} \right\|} ^2},
\end{array}
\end{equation}
where $l$ is used to represent loss function, MSE. And $p$ stands for the set of parameters in ESRGCNN model.
\subsection{Group enhanced convolutional block}
A 6-layer group enhanced convolutional block is used to strength representation ability of low-frequency features for boosting performance and speed in SISR, according to correlations of different channels. Also, taking network design principle into consideration, a signal enhancement idea is used to inherit long-distance features for handling poorer memory ability problem of shallow layers in a deep network. Detailed information of the GEB is listed as follows.

It is known that some of existing SR networks directly enhanced deep hierarchical features rather than fully using correlation information of different channels to improve SISR performance, which had higher computational cost. Taking this factor above into account, we use group convolutions to split obtained features from first four convolutional layers in each GEB as two groups: GConv1 known a distilling part (Quarter of channel number of obtained features from current convolutional layer) and GConv2 (Three quarters of channel number of obtained features from current convolutional layer) known a remaining part, as shown in Fig.1. In this paper, the GConv1 is a group convolution of input channel number of 16, output channel number of 16 and filter size of $3 \times 3$. The GConv2 is a group convolution of input channel number of 48, output channel number of 48 and filter size of $3 \times 3$. Specifically, only chosen the remaining part as the input of next convolutional layer in main network is used to extract more deep features, where can boost the training  efficiency of a SR model. Also, differ from most of SR methods, only fusing two adjacent GConv2 in the GEB can enhance the effect of deep neighborhood information for improving the expressive ability of low-frequency features. Besides, the distilling parts besides the first distilling part indirectly rely on the previous remaining parts and fusing these obtained features can offer complementary wide information for the remaining part of deep layer to improve the SISR performance. To vividly describe the mentioned implementations of each GConv1 and GConv2, we formulated the process above as follows.

Simultaneously learning these features of each groups and fusing them to strengthen their connections for improving expressive ability of low-frequency features as the following steps.

First step fuses obtained features of two adjacent GConv2 via the RL technique as the input of next convolutional layer to facilitate more accurate deep features of from different channels in terms of enhancing features of adjacent layers in SISR. Specifically, the outputs of the first GConv2 from different GEBs can be represented as
\begin{small}
\begin{equation}
   O\_GConv2_j^{1} = \left\{ {\begin{array}{*{1}{l}}
\frac{{3s}}{4}(C(R(C(I_{LR})))), j = 1\\
\frac{{3s}}{4}(C(O\_GE{B_{j - 1}})),j=2,3,4,5,6,
\end{array}} \right.
\end{equation}
\end{small}
where $O\_GE{B_{j - 1}}$ denotes output of the $j-1th$ GEB. Also, $O\_GConv2_j^1$ is output of the first GConv2 from the $jth$ GEB. $\frac{{3s}}{4}$ expresses remaining channel number of a convolutional layer, where s is 64. Additionally, outputs of other GConv2 from different GEBs can be  expressed as:
\begin{small}
\begin{equation}
O\_GConv2_j^i = O\_GConv2_j^{i - 1} + \frac{{3s}}{4}(C(R(O\_GConv2_j^{i - 1}))),
\end{equation}
\end{small}
where $O\_GConv2_j^i$ stands for the output of the $ith$ GConv2 in the $jth$ GEB. Also, $i = 2,3,4$ and $j = 1,2,3,4,5,6$. Specifically, plus denotes the RL operation, which is also expressed as $\oplus$ in Fig.1.

Differ from the GConv2, output of the first GConv1 in the GEB can be obtained by the last quarter of output channel from the first convolutional layer. Outputs of other GConv1 in the GEB are obtained by upper GConv2. That is, the upper GConv2 connects a ReLU to convert linear features into non-linearity. Then, the non-linearity acts as a convolutional layer to further learn low-frequency features. Last quarter channels of output information of obtained features from the mentioned convolutional layer are as input of GConv1. That can be shown as Eqs.(5) and (6).

\begin{small}
\begin{equation}
   O\_GConv1_j^{1} = \left\{ {\begin{array}{*{1}{l}}
\frac{{s}}{4}(C(R(C(I_{LR})))), j = 1\\
\frac{{s}}{4}(C(O\_GE{B_{j - 1}})),j=2,3,4,5,6,
\end{array}} \right.
\end{equation}
\end{small}
\begin{small}
\begin{equation}
  O\_GConv1_j^i = \frac{s}{4}(C(R(O\_GConv2_{j - 1}^i))),
\end{equation}
\end{small}
where $O\_GConv1_j^i$ and $O\_GConv1_j^i$ represent the outputs of the $ith$ GConv1 and  $1-th$ GConv1 from the $jth$ GEB, respectively.

Second step uses a RL technique to merge obtained features from all the GConv1 for strengthening the connection of different distilling parts as follows.
\begin{small}
\begin{equation}
  O\_TGConv1_j^4 = \sum\limits_{i = 1}^4 {O\_GConv1_j^i},
\end{equation}
\end{small}
where $O\_TGConv{1_j}$ is output of all the GConv1 of the $jth$ GEB .

Third step utilizes concatenation operation to integrate obtained feature from the last GConv1 and GConv2 in terms of channels for obtaining more complementary features, which uses Eq. (8) to express the process above.

\begin{small}
\begin{equation}
  O\_E{F_j} = R(Concat(O\_GConv2_j^4,O\_TGConv1_j^4)),
\end{equation}
\end{small}
where $O\_E{F_j}$ is the output of enhanced features from the $jth$ GEB. Its output channel number is 64. Also, $Concat$ denotes a concatenation operation in Eq.(8) and Fig.1. Additionally, $O\_E{F_j}$ acts two layers of Conv+ReLU, where Conv+ReLU expresses a convolutional layer connects an activation function of ReLU. Also, input channel number, output channel number, filter size of each layer are 64, 64 and $3 \times 3$, respectively. That can be formulated as

\begin{small}
\begin{equation}
  O\_D{F_j} = R(C(R(C(O\_E{F_j})))),
\end{equation}
\end{small}
where $O\_D{F_j}$ is the output of two layers of Conv+ReLU in the $jth$ GEB and $j=1,2,3,4,5,6$.

Due to deeper network architecture, effect of shallow layers on deep layers may weaken. Inspired by that, we use a signal enhancement idea to inherit long-distance features for handling memory ability problem of shallow layers for the whole network. That is, obtained features of a shallow layer are overlaid on obtained features of a deep layer through a RL technique to improve the importance of shallow layers in SISR. The mentioned illustrations can be shown as Eq.(10).

\begin{small}
\begin{equation}
O\_GE{B_j} = O\_D{F_j} + O\_GE{B_{j - 1}},
\end{equation}
\end{small}
where $O\_GE{B_j}$ stands for output of the $jth$ GEB and $j=2,3,4,5,6$. Specifically,                           
$O\_GE{B_1}{\rm{ = }}O\_D{F_1} + R(C({I_{LR}}))$. Additionally, the $O\_GE{B_6}$ acts a Conv+ReLU as shown in Eq.(11), which is used to prevent transition enhancement of Eq.(10).

\begin{small}
\begin{equation}
{O_{CR}} = R(C(O\_GE{B_6})),
\end{equation}
\end{small}
where ${O_{CR}}$ denotes output of the last Conv+ReLU and is input of the upsampling layer, which is given in Section 3.4.
\subsection{Adaptive upsampling mechanism}
It is known that some popular SR methods use a model for single certain scale.
However, LR images in practice are unknown corruption \cite{agustsson2017ntire}. Motivated by that, we use an adaptive upsampling mechanism with a flexible valve \cite{tian2020lightweight,ahn2018fast} to achieve high-quality images from LR images of different scales for achieving a flexible SR model as follows.

The adaptive upsampling mechanism is composed of three modes from different scales (i.e., $\times 2$, $\times 3$ and $\times 4$). The mentioned three modes include one Conv+Shuffle $\times 2$ (regarded as mode  $\times 2$), one Conv+Shuffle  $\times 3$ (regarded as  $\times 3$ mode) and two Conv+Shuffle  $\times 2$  (regarded as  $\times 4$ mode), where Conv+Shuffle  $\times 2$  and Conv+Shuffle  $\times 3$ express a convolution of $3\times 3$ acts a Shuffle $\times 2$ and Shuffle $\times 3$, respectively. The  $\times 4$ mode is composed of two stacked $\times 2$ modes. Also, input channel number of input and output from these convolutions are 64. Specifically, these modes rely on a valve to implement a blind SR model. When the valve is set as 0, three modes work in parallel to train a SR model for $\times 2$, $\times 3$ and $\times 4$  as shown in Fig.2. When the valve is only set as one of among 2, 3 and 4, one of three modes is chosen to test a blind SR model in Fig.3. Further, the function of upsampling technique can use Eq. (12) to stand for the mentioned process. 

\begin{small}
\begin{equation}
   O_{UP} = \left\{ {\begin{array}{*{1}{l}}
UP({O_{CR}})\\
{S_2}{\rm{(}}C(O_{CR}^2)) \odot {S_3}{\rm{(}}C(O_{CR}^3)) \odot {S_2}(C({S_2}{\rm{(}}C(O_{CR}^4)))),
\end{array}} \right.
\end{equation}
\end{small}
where ${O_{UP}}$ expresses output of the upsampling layer. ${S_2}$, ${S_3}$ and ${S_4}$ represent the functions of Shuffle $\times 2$, Shuffle $\times 3$ and Shuffle $\times 4$, respectively. $O_{CR}^2$, $O_{CR}^3$ and $O_{CR}^4$ are outputs of obtained low-frequency information from $\times 2$,  $\times 3$ and $\times 4$, respectively. Also, let $\odot$  stands for a splicing operation of tensor in Pytorch. Besides, $O_{UP}$ is input of the last convolutional layer in Eq. (13), which is used to construct the predicted SR image. Also, its input number, output number and filter size are 64, 3 and $3 \times 3$, respectively.

\begin{equation}
{I_{SR}} = C({O_{UP}}).
\end{equation}

\begin{figure}[!htbp]
\centering
\subfloat{\includegraphics[width=3.5in]{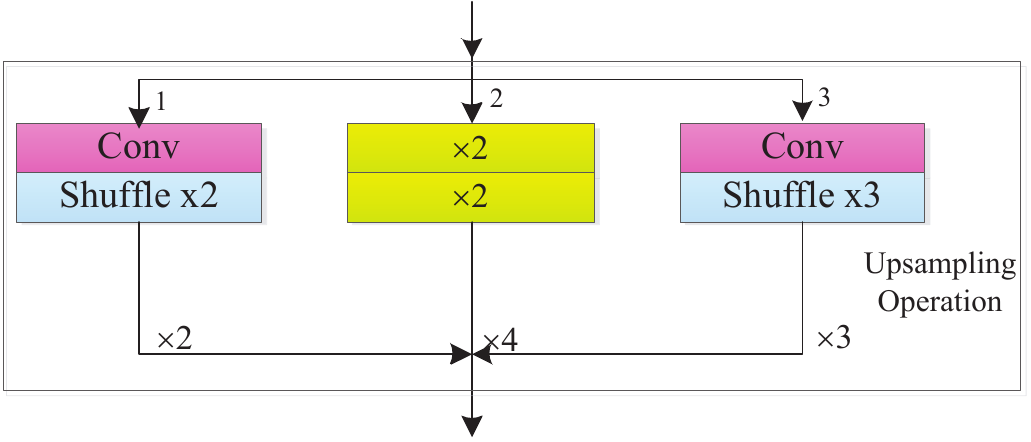}
\label{fig_second_case}}
\caption{A parallel upsampling operation for training a blind SR model.}
\label{fig:7}
\end{figure}

\begin{figure}[!htbp]
\centering
\subfloat{\includegraphics[width=3.5in]{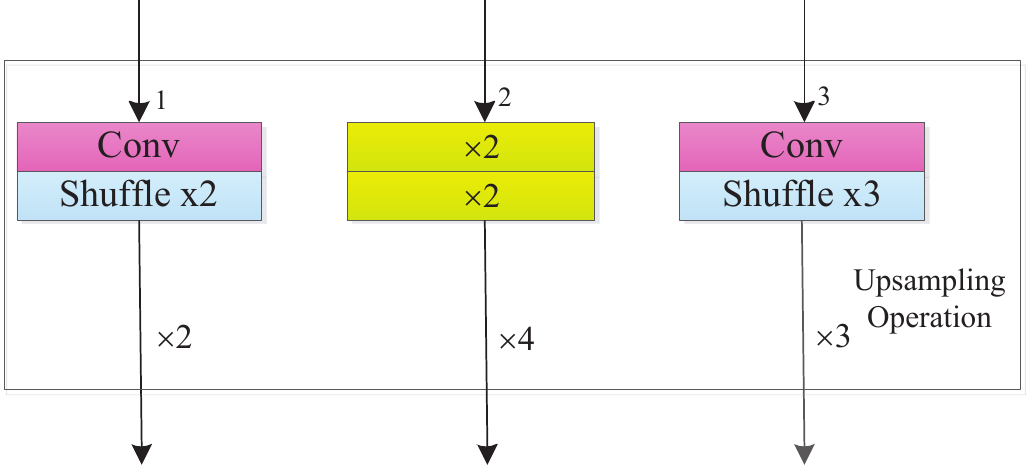}
\label{fig_second_case}}
\caption{A upsampling operation for testing a blind SR model.}
\label{fig:7}
\end{figure}

\section{Experiments and analysis}
\subsection{Training dataset}
Following popular SR methods in \cite{tian2020coarse,tian2020lightweight,ahn2018fast}, the known SR dataset, DIV2K \cite{agustsson2017ntire} is chosen to train a ESRGCNN model. Specifically, the DIV2K dataset is composed of 800 training color images, 100 validation color images and 100 test color images from three scales (i.e., $\times 2$, $\times 3$, and $\times 4$). Besides, due to shortage of training samples, obtained SR model is not robust. To prevent this phenomenon, we utilize the following data augment way \cite{tian2020coarse} to expand the training dataset. First step incorporates color images of the same scale from training dataset and validation dataset to make up of a new training dataset of ESRGCNN model. Also, taking efficiency of ESRGCNN into account, we crop each given LR image from this new training dataset as patches of $83 \times 83$. Second step exploits random horizontal flips and rotation operation of $90^\circ$ to enhance mentioned patches.
\subsection{Test datasets}
To fairly test the performance of ESRGCNN in SISR, four benchmark datasets, including Set5 \cite{bevilacqua2012low}, Set14 \cite{bevilacqua2012low}, BSD100 \cite{martin2001database} and Urban100 \cite{huang2015single} of three scales (i.e.,$\times 2$, $\times 3$, and $\times 4$) are used as test datasets in this paper. Specifically, the Set5 and Set14 are composed of five and fourteen color nature images of three different scales, respectively. The BSD100 (abbreviated as B100) and Urban100 (abbreviated as U100) contain 100 different color images with $\times 2$, $\times 3$, and $\times 4$, respectively.

Following the state-of-the-art SR methods, such as MemNet \cite{mao2016image} and LESRCNN \cite{tian2020lightweight}, Y channel of YCbCr space is used to design comparative experiments in this paper. That is, the obtained RGB images from the ESRGCNN are required to transform as the Y channel images to valid the performance of image super-resolution.
\subsection{Experimental Settings}
The following initial parameters are utilized to train a SR model of ESRGCNN. Specifically, original learning rate is 1e-4, which halves every 4e+5 steps in all the 6e+5 steps. Also, batch size of 32, epsilon of 1e-8, beta\_1 of 0.9, beta\_2 of 0.999 and other initial parameters of Refs. \cite{tian2020coarse,tian2020lightweight,ahn2018fast} are used to learn ESRGCNN model. Besides, parameters of ESRGCNN in the training process are updated by optimizer of Adam \cite{kingma2014adam}.

The ESRGCNN model is conducted through Pytorch of 0.41 and Python of 2.7 on Ubuntu of 16.04. Besides, the supplementary PC with RAM of 16G is composed of one GPU of Inter Core i7-7800 and two GPUs of Nvidia GeForce GTX 1080Ti, where these GPUs can accelerate the running speed by Nvidia CUDA of 9.0 and cuDNN of 7.5.
\subsection{Network analysis}
It is known that diversity of expanding network can boost the expressive ability of deep network \cite{tian2020image}. Also, most of existing SR methods enhance the SR performance via treating all information of different channels, which may cause huge computational cost. Inspired by that, we propose an enhanced super-resolution group CNN (ESRGCNN) by splitting channels to expand the network width for enhancing expressive ability of low-frequency information in SISR. A signal enhancement operation in this ESRGCNN is used to transfer long-distance contextual information to solve long-term dependency problem, according to signal process idea and network design principle. Additionally, an adaptive up-sampling mechanism is fused into the ESRGCNN to train a SR model. More details of network design principle are shown as follows.

ESRGCNN is composed of two Conv+ReLU, several group enhanced convolutional blocks (GEBs), an up-sampling operation mechanism and a Conv. Specifically, the first Conv+ReLU is used to covert LR image into non-linear low-frequency features. It is known that deep layer can learn more accurate features, according to VGG architecture \cite{simonyan2014very}. Motived by that, the first Conv+ReLU acts six stacked GEBs to mine more robust low-frequency features. The design of each GEB breaks the rules of promoting SR performance and reducing complexity.

In terms of SR performance, a GEB uses correlation information of different channels to obtain more accurate low-frequency features as follows. Firstly, first four convolutional layers in each GEB can be split as two groups via group convolutions: GConv1 and GConv2. The GConv1 (also called distilling part) and GConv2 (also called remaining part) take quarter and three quarters of channel number of obtained features from current convolutional layer, respectively. To extract richer deep features, we only choose GConv2 as input of next convolutional layer in main network. Also, it can improve the training speed of SR model and reduce the complexity of SR network. Additionally, most of existing SR networks only fuse hierarchical features of each layer or width features of different sub-networks to improve the SR effect. For example, some of existing SR methods only use current layer to act each latter layer in deep network for addressing long-term dependency problem of deep network and promoting SR performance. However, this method may increase execution time in recovering high-quality image. Taking design principle of deep network and efficiency in SISR into account, we combine deep feature enhancement and wide feature fusion way to address problem above.

For the deep feature enhancement, we present a two-step mechanism to mine more accurate low-frequency features. First step only fuses two adjacent GConv2 via residual learning operation to enhance the correlation of deep neighborhood context for improving the expressive ability of low-frequency features, where its effectiveness is proved as illustrated in Table 1. That is, ESRGCNN without last CR and wide feature fusion (WFF) is superior to ESRGCNN without the last CR, WFF and distilling parts in Peak signal-to-noise ratio (PSNR) \cite{hore2010image} and structural similarity index (SSIM) \cite{hore2010image} on U100 for $\times 2$ upscaling, where CR denotes Conv+ReLU.
\begin{table}[htbp!]
\caption{PSNR and SSIM of different SR methods on U100 for $\times 2$.}
\label{tab:1}
\centering
\scalebox{0.7}[0.75]{
\begin{tabular}{|c|c|c|}
\hline
Methods &PSNR (dB) &SSIM\\
\hline
ESRGCNN without last CR, WFF, distilling parts and remaining parts  &12.86	&0.3788\\
\hline
ESRGCNN without the last CR, WFF and distilling parts	&31.20 &0.9182\\
\hline
ESRGCNN without the last CR, WFF and group convolutions &31.23 &0.9180\\
\hline
ESRGCNN without last CR and wide feature fusion (WFF) &31.31 &0.9192\\
\hline
ESRGCNN without the last Conv+ReLU (CR) &\textcolor{blue}{31.97} &\textcolor{blue}{0.9266}\\
\hline
ESRGCNN	&\textcolor{red}{32.02} &\textcolor{red}{0.9268}\\
\hline
\end{tabular}}
\label{tab:booktabs}
\end{table}

Second step merges obtained features of each GConv1 through residual operation to enhance the effect of deep hierarchical channel features. Specifically, GConv1 besides the first GConv1 indirectly by the previous GConv2, which can be shown in details in Section 3.3. Thus, the second step is complementary with the first step, which can be introduced in wide feature fusion. The mentioned fact is tested that ESRGCNN without the last CR, WFF and distilling parts outperforms ESRGCNN without last CR, WFF, distilling parts and remaining parts on U100 for $\times 2$ upscaling as shown in Table 1. According to the previous section, it is known that ESRGCNN uses correlations of different channels to strengthen effect of part channels rather than full channels for improving SR performance, where its effectiveness and complexity are shown in Tables 2 and 3. We can see that ESRGCNN without the last CR and WFF outperforms ESRGCNN without the last CR, WFF and group convolutions in both of PSNR and SSIM of recovering high-quality images of different sizes in Table 2, and complexity, i.e., parameters and flops in Table 3. Specifically, red and blue lines
are defined as the highest and second PSNR and SSIM values  in Tables 1, 2 and 3, respectively.
\begin{table}[htbp!]
\caption{ PSNR and SSIM of different SR methods for recovering high-quality images with different sizes.}
\label{tab:1}
\centering
\scalebox{0.7}[0.7]{
\begin{tabular}{|c|c|c|c|c|}
\hline
\multirow{2}{*}{Methods}
&$256 \times 256$ &$512\times 512$  \\
\cline{2-3} &PSNR(dB)/SSIM	&PSNR(dB)/SSIM\\
\hline
ESRGCNN without the last CR, WFF and group convolutions	&\textcolor{red}{28.90/}\textcolor{blue}{0.8576}	&\textcolor{blue}{30.41/0.8943}\\
\hline
ESRGCNN without the last CR and WFF	&\textcolor{red}{28.90/0.8584}  &\textcolor{red}{30.44/0.8952}\\
\hline
\end{tabular}}
\label{tab:booktabs}
\end{table}

\begin{table}[htbp!]
\caption{Complexity of different SR networks.}
\label{tab:1}
\centering
\scalebox{0.80}[0.85]{
\begin{tabular}{|c|c|c|}
\hline
Methods &Parameters &Flops\\
\hline
ESRGCNN without the last CR, WFF and group convolutions  &\textcolor{blue}{1,367K}	&\textcolor{blue}{10.21G}\\
\hline
ESRGCNN without the last CR and WFF	&\textcolor{red}{1,202K}	&\textcolor{red}{9.08G}\\
\hline
\end{tabular}}
\label{tab:booktabs}
\end{table}

\begin{table}[t!]
\caption{Average PSNR/SSIM results of different SR methods for three different upscaling ($\times 2$, $\times 3$ and $\times 4$) on Set5.}
\label{tab:1}
\centering
\scalebox{0.80}[0.75]{
\begin{tabular}{|c|c|c|c|c|}
\hline
\multirow{2}{*}{Dataset} &
\multirow{2}{*}{Methods} &
$\times 2$ & $\times 3$ & $\times 4$\\
\cline{3-5} & &PSNR/SSIM &PSNR/SSIM &PSNR/SSIM\\
\hline
\multirow{34}{*}{Set5} &
Bicubic	&33.66/0.9299	&30.39/0.8682	&28.42/0.8104\\
\cline{2-5} &
A+\cite{timofte2014a+}	&36.54/0.9544	&32.58/0.9088	&30.28/0.8603\\
\cline{2-5} &
RFL \cite{schulter2015fast}	&36.54/0.9537	&32.43/0.9057	&30.14/0.8548\\
\cline{2-5} &
SelfEx\cite{huang2015single}	&36.49/0.9537 &32.58/0.9093 &30.31/0.8619\\
\cline{2-5} &
CSCN\cite{wang2015deep} &36.93/0.9552 &33.10/0.9144 &30.86/0.8732\\
\cline{2-5} &
RED30\cite{mao2016image} &37.56/0.9595 &33.70/0.9222	&31.33/0.8847\\
\cline{2-5} &
DnCNN\cite{zhang2017beyond}	&37.58/0.9590	&33.75/0.9222 &31.40/0.8845\\
\cline{2-5} &
TNRD\cite{chen2016trainable} &36.86/0.9556	&33.18/0.9152 &30.85/0.8732\\
\cline{2-5} &
FDSR\cite{lu2018fast} &37.40/0.9513	&33.68/0.9096 &31.28/0.8658\\
\cline{2-5} &
SRCNN\cite{dong2015image}	&36.66/0.9542	&32.75/0.9090 &30.48/0.8628\\
\cline{2-5} &
FSRCNN\cite{dong2016accelerating} &37.00/0.9558	&33.16/0.9140 &30.71/0.8657\\
\cline{2-5} &
RCN\cite{shi2017structure} &37.17/0.9583	&33.45/0.9175	&31.11/0.8736\\
\cline{2-5} &
VDSR\cite{kim2016accurate} &37.53/0.9587	&33.66/0.9213	&31.35/0.8838\\
\cline{2-5} &
DRCN\cite{kim2016deeply} &37.63/0.9588	&33.82/0.9226	&31.53/0.8854\\
\cline{2-5} &
CNF\cite{ren2017image}	&37.66/0.9590	&33.74/0.9226	&31.55/0.8856\\
\cline{2-5} &
LapSRN\cite{lai2017deep} &37.52/0.9590	&-	&31.54/0.8850\\
\cline{2-5} &
IDN\cite{hui2018fast}  &\textcolor{blue}{37.83}/\textcolor{blue}{0.9600} &34.11/\textcolor{red}{0.9253} &31.82/0.8903\\
\cline{2-5} &
DRRN\cite{tai2017image} &37.74/0.9591  &34.03/0.9244 &31.68/0.8888 \\
\cline{2-5} &
BTSRN\cite{fan2017balanced} &37.75/- &34.03/- &31.85/-\\
\cline{2-5} &
MemNet\cite{tai2017memnet}  &37.78/0.9597  &34.09/0.9248 &31.74/0.8893\\
\cline{2-5} &
CARN-M\cite{ahn2018fast}  &37.53/0.9583 &33.99/0.9236 &31.92/0.8903\\
\cline{2-5} &
EEDS+\cite{wang2019end} &37.78/\textcolor{red}{0.9609} &33.81/\textcolor{blue}{0.9252} &31.53/0.8869\\
\cline{2-5} &
DRFN\cite{yang2018drfn} &37.71/0.9595 &34.01/0.9234 &31.55/0.8861\\
\cline{2-5} &
MADNet-${L_1}$\cite{lan2020madnet} &\textcolor{red}{37.85}/\textcolor{blue}{0.9600} &\textcolor{blue}{34.16}/\textcolor{red}{0.9253} &31.95/0.8917\\
\cline{2-5} &
MSDEPC\cite{liu2019single} &37.39/0.9576 &33.37/0.9184 &31.05/0.8797\\
\cline{2-5} &
MADNet-${L_F}$\cite{lan2020madnet} &\textcolor{red}{37.85}/\textcolor{blue}{0.9600} &34.14/0.9251 &\textcolor{blue}{32.01}/\textcolor{red}{0.8925}\\
\cline{2-5}&
LESRCNN\cite{tian2020lightweight} 	&37.57/0.9582	&34.05/0.9238	&31.88/0.8907\\
\cline{2-5} &
DIP-FKP\cite{liang2021flow}  &30.16/0.8637 &28.82/0.8202 &27.77/0.7914\\
\cline{2-5} &
DIP-FKP+USRNet\cite{liang2021flow} &32.34/0.9308 &30.78/0.8840 &29.29/0.8508\\
\cline{2-5} &
KOALAnet\cite{kim2021koalanet} &33.08/0.9137 &- &30.28/0.8658\\
\cline{2-5} &
FALSR-C\cite{chu2021fast} & 37.66/0.9586 &- &-\\
\cline{2-5} &
SRCondenseNet\cite{jain2018efficient} &37.79/0.9594 &- &-\\
\cline{2-5} &
SPSR\cite{ma2020structure} 
&30.40/0.8627 &- &-\\
\cline{2-5} &
DWSR\cite{guo2017deep}
&37.43/0.9568 &33.82/0.9215 &31.39/0.8833\\
\cline{2-5} &
S-BayeSR \cite{gao2022bayesian} &31.50/0.8805 &- &-\\
\cline{2-5} &
ESRGCNN (Ours)	&37.79/0.9589	&\textcolor{red}{34.24}/\textcolor{blue}{0.9252}	&\textcolor{red}{32.02}/\textcolor{blue}{0.8920}\\
\hline
\end{tabular}}
\label{tab:booktabs}
\end{table}

\begin{table}[t!]
\caption{Average PSNR/SSIM results of different SR methods for three different upscaling ($\times 2$, $\times 3$ and $\times 4$) on Set14.}
\label{tab:1}
\centering
\scalebox{0.90}[0.85]{
\begin{tabular}{|c|c|c|c|c|}
\hline
\multirow{2}{*}{Dataset} &
\multirow{2}{*}{Methods} &
$\times 2$ & $\times 3$ & $\times 4$\\
\cline{3-5} & &PSNR/SSIM &PSNR/SSIM &PSNR/SSIM\\
\hline
\multirow{34}{*}{Set14} &
Bicubic	&30.24/0.8688	&27.55/0.7742	&26.00/0.7027\\
\cline{2-5} &
A+\cite{timofte2014a+}	&32.28/0.9056	&29.13/0.8188	&27.32/0.7491\\
\cline{2-5} &
RFL\cite{schulter2015fast}	&32.26/0.9040	&29.05/0.8164	&27.24/0.7451\\
\cline{2-5} &
SelfEx\cite{huang2015single}	&32.22/0.9034 &29.16/0.8196	&27.40/0.7518\\
\cline{2-5} &
CSCN\cite{wang2015deep} &32.56/0.9074	&29.41/0.8238	&27.64/0.7578\\
\cline{2-5} &
RED30 \cite{mao2016image} &32.94/0.9144	&29.61/0.8341	&27.86/0.7718\\
\cline{2-5} &
DnCNN\cite{zhang2017beyond} &33.03/0.9128	&29.81/0.8321	&28.04/0.7672\\
\cline{2-5} &
TNRD\cite{chen2016trainable} &32.51/0.9069	&29.43/0.8232	&27.66/0.7563\\
\cline{2-5} &
FDSR\cite{lu2018fast} &33.00/0.9042	&29.61/0.8179	&27.86/0.7500\\
\cline{2-5} &
SRCNN\cite{dong2015image} &32.42/0.9063 &29.28/0.8209 &27.49/0.7503\\
\cline{2-5} &
FSRCNN\cite{dong2016accelerating} &32.63/0.9088 &29.43/0.8242	&27.59/0.7535\\
\cline{2-5} &
RCN\cite{shi2017structure} &32.77/0.9109	&29.63/0.8269	&27.79/0.7594\\
\cline{2-5} &
VDSR\cite{kim2016accurate} &33.03/0.9124	&29.77/0.8314	&28.01/0.7674\\
\cline{2-5} &
DRCN\cite{kim2016deeply} &33.04/0.9118	&29.76/0.8311	&28.02/0.7670\\
\cline{2-5} &
CNF\cite{ren2017image} &33.38/0.9136 &29.90/0.8322 &28.15/0.7680\\
\cline{2-5} &
LapSRN\cite{lai2017deep} &33.08/0.9130	&29.63/0.8269 &28.19/0.7720\\
\cline{2-5} &
IDN\cite{hui2018fast} &33.30/0.9148 &29.99/0.8354 &28.25/0.7730\\
\cline{2-5} &
DRRN\cite{tai2017image} &33.23/0.9136	&29.96/0.8349	&28.21/0.7720\\
\cline{2-5} &
BTSRN\cite{fan2017balanced} &33.20/- &29.90/- &28.20/-\\
\cline{2-5} &
MemNet\cite{tai2017memnet}	&33.28/0.9142	&30.00/0.8350	&28.26/0.7723\\
\cline{2-5} &
CARN-M\cite{ahn2018fast} &33.26/0.9141	&30.08/0.8367	&28.42/0.7762\\
\cline{2-5} &
EEDS+\cite{wang2019end} &33.21/0.9151 &29.85/0.8339 &28.13/0.7698\\
\cline{2-5} &
DRFN\cite{yang2018drfn} &33.29/0.9142 &30.06/0.8366 &28.30/0.7737\\
\cline{2-5} &
MADNet-${L_1}$\cite{lan2020madnet}	&33.38/\textcolor{blue}{0.9161} &\textcolor{blue}{30.21/0.8398}	&28.44/0.7780\\
\cline{2-5} &
MSDEPC\cite{liu2019single} &32.94/0.9111	&29.62/0.8279	&27.79/0.7581\\
\cline{2-5} &
MADNet-${L_F}$\cite{lan2020madnet} &\textcolor{blue}{33.39/0.9161} &30.20/0.8395 &\textcolor{blue}{28.45/0.7781}\\
\cline{2-5} &
LESRCNN\cite{tian2020lightweight}	&33.30/0.9145	&30.16/0.8384	&28.43/0.7776\\
\cline{2-5} &
DIP-FKP\cite{liang2021flow}  &27.06/0.7421 &26.27/0.6922 &25.65/0.6764\\
\cline{2-5} &
DIP-FKP+USRNet\cite{liang2021flow}  &28.18/0.8088 & 27.76/0.7750 &26.70/0.7383\\
\cline{2-5} &
KOALAnet\cite{kim2021koalanet} &30.35/0.8568 &- &27.20/0.7541\\
\cline{2-5} &
FALSR-C\cite{chu2021fast} &  33.26/0.9140 &- &-\\
\cline{2-5} &
SRCondenseNet\cite{jain2018efficient} &33.23/0.9137 &- &-\\
\cline{2-5} &
SPSR\cite{ma2020structure} 
&26.64/0.7930 &- &-\\
\cline{2-5} &
DWSR\cite{guo2017deep}
&33.07/0.9106 &29.83/0.8308 &28.04/0.7669\\
\cline{2-5} &
S-BayeSR \cite{gao2022bayesian} &28.08/0.7561 &- &-\\
\cline{2-5} &
ESRGCNN (Ours)	&\textcolor{red}{33.48/0.9166}	&\textcolor{red}{30.29/0.8413}	&\textcolor{red}{28.57/0.7801}\\
\hline
\end{tabular}}
\label{tab:booktabs}
\end{table}

It is known that different views can extract diverse information for image processing applications \cite{zhang2018image}. Inspired by that, we fuse features from different branches of one network to obtain more complementary low-frequency features in Eq.(8). That is, using a concatenation operation fuses wide features of the first and second steps to obtain more robust features for SISR, where the effectiveness of wide feature fusion is proved by ESRGCNN without last CR and wide feature fusion (WFF) and ESRGCNN without the last CR, WFF and distilling parts in Table 1.

To prevent obtained features with redundant information of the mentioned operation, a two-layer stacked convolutional layers are used to extract more accurate low-frequency features. Additionally, due to deeper network architecture, memory ability of shallow layers gets poorer on the whole network. Motivated by that, a signal enhancement idea is presented to extract long-distance features for resolving long-term dependency problem in deep network. That is, this signal enhancement is implemented by using a RL technique to merge input and output of GEB as the whole output of GEB, where its good result is validated ESRGCNN without the last Conv+ReLU (CR) and ESRGCNN without last CR and wide feature fusion (WFF) as illustrated in Table 1. To make obtained low-frequency features from six GEBs smoother, we choose a CR to extract more accurate low-frequency features. The good performance of last CR is tested by ESRGCNN and ESRGCNN without the last CR in Table 1. Subsequently, to deal with varying scales, an adaptive upsampling operation with a flexible valve in Eq.(12) is presented to train a blind model. Also, it can transform obtained low-frequency features into high-frequency features. It acts a convolutional layer, which is used to construct a high-quality image.

\begin{table}[t!]
\caption{Average PSNR/SSIM results of different SR methods for three different upscaling ($\times 2$, $\times 3$ and $\times 4$) on B100.}
\label{tab:1}
\centering
\scalebox{0.90}[0.78]{
\begin{tabular}{|c|c|c|c|c|}
\hline
\multirow{2}{*}{Dataset} &
\multirow{2}{*}{Methods} &
$\times 2$ & $\times 3$ & $\times 4$\\
\cline{3-5} & &PSNR/SSIM &PSNR/SSIM &PSNR/SSIM\\
\hline
\multirow{32}{*}{B100} &
Bicubic	&29.56/0.8431	&27.21/0.7385	&25.96/0.6675\\
\cline{2-5} &
A+\cite{timofte2014a+}	&31.21/0.8863	&28.29/0.7835	&26.82/0.7087\\
\cline{2-5} &
RFL\cite{schulter2015fast} &31.16/0.8840	&28.22/0.7806	&26.75/0.7054\\
\cline{2-5} &
SelfEx\cite{huang2015single}	&31.18/0.8855	&28.29/0.7840 &26.84/0.7106\\
\cline{2-5} &
CSCN\cite{wang2015deep} &31.40/0.8884	&28.50/0.7885 &27.03/0.7161\\
\cline{2-5} &
RED30\cite{mao2016image} &31.98/0.8974	&28.92/0.7993 &27.39/0.7286\\
\cline{2-5} &
DnCNN\cite{zhang2017beyond}	&31.90/0.8961	&28.85/0.7981	&27.29/0.7253\\
\cline{2-5} &
TNRD\cite{chen2016trainable} &31.40/0.8878	&28.50/0.7881	&27.00/0.7140\\
\cline{2-5} &
FDSR\cite{lu2018fast} &31.87/0.8847	&28.82/0.7797	&27.31/0.7031\\
\cline{2-5} &
SRCNN\cite{dong2015image}	&31.36/0.8879	&28.41/0.7863	&26.90/0.7101\\
\cline{2-5} &
FSRCNN\cite{dong2016accelerating} &31.53/0.8920	&28.53/0.7910	&26.98/0.7150\\
\cline{2-5} &
VDSR\cite{kim2016accurate} &31.90/0.8960	&28.82/0.7976	&27.29/0.7251\\
\cline{2-5} &
DRCN\cite{kim2016deeply} &31.85/0.8942	&28.80/0.7963	&27.23/0.7233\\
\cline{2-5} &
CNF\cite{ren2017image}	&31.91/0.8962	&28.82/0.7980	&27.32/0.7253\\
\cline{2-5} &
LapSRN\cite{lai2017deep}	&31.80/0.8950 &-	&27.32/0.7280\\
\cline{2-5} &
IDN\cite{hui2018fast} &\textcolor{red}{32.08}/\textcolor{red}{0.8985} &28.95/0.8013 &27.41/0.7297\\
\cline{2-5} &
DRRN\cite{tai2017image} &32.05/0.8973 &28.95/0.8004 &27.38/0.7284\\
\cline{2-5} &
BTSRN\cite{fan2017balanced} &32.05/-    &28.97/- &27.47/- \\
\cline{2-5} &
MemNet\cite{tai2017memnet} &32.08/0.8978 &28.96/0.8001 &27.40/0.7281\\
\cline{2-5} &
CARN-M\cite{ahn2018fast}	&31.92/0.8960	&28.91/0.8000	&27.44/0.7304\\
\cline{2-5} &
EEDS+\cite{wang2019end} &31.95/0.8963 &28.88/\textcolor{red}{0.8054} &27.35/0.7263\\
\cline{2-5} &
DRFN\cite{yang2018drfn} &32.02/\textcolor{blue}{0.8979} &28.93/0.8010  &27.39/0.7293\\
\cline{2-5} &
MADNet-${L_1}$\cite{lan2020madnet} &32.04/\textcolor{blue}{0.8979} &\textcolor{blue}{28.98}/0.8023 &\textcolor{blue}{27.47/0.7327}\\
\cline{2-5} &
MSDEPC\cite{liu2019single} &31.64/0.8961 &28.58/0.7918	&27.10/0.7193\\
\cline{2-5} &
MADNet-${L_F}$\cite{lan2020madnet}  &\textcolor{blue}{32.05}/0.8981 &\textcolor{blue}{28.98}/0.8023 &\textcolor{blue}{27.47/0.7327}\\
\cline{2-5} &
LESRCNN\cite{tian2020lightweight}	&31.95/0.8964	&28.94/0.8012	&\textcolor{blue}{27.47}/0.7321\\
\cline{2-5} &
DIP-FKP\cite{liang2021flow}  &26.72/0.7089 & 25.96/0.6660 & 25.15/0.6354\\
\cline{2-5} &
DIP-FKP+USRNet\cite{liang2021flow}  & 28.61/0.8206 &27.29/0.7484 & 25.97/0.6902\\
\cline{2-5} &
KOALAnet\cite{kim2021koalanet} &29.70/0.8248 &- &26.97/0.7172\\
\cline{2-5} &
FALSR-C\cite{chu2021fast} &  31.96/0.8965 &- &-\\
\cline{2-5} &
SPSR\cite{ma2020structure} 
&25.51/0.6576 &- &-\\
\cline{2-5} &
DWSR\cite{guo2017deep}
&31.80/0.8940 &- &27.25/0.7240\\
\cline{2-5} &S-BayeSR \cite{gao2022bayesian}
&27.21/0.7091
&- &-\\
\cline{2-5} &
ESRGCNN (Ours)	&\textcolor{red}{32.08}/0.8978	&\textcolor{red}{29.05}/\textcolor{blue}{0.8036}	&\textcolor{red}{27.57/0.7348}\\
\hline
\end{tabular}}
\label{tab:booktabs}
\end{table}
\begin{table}[t!]
\caption{Average PSNR/SSIM results of different SR methods for three different upscaling ($\times 2$, $\times 3$ and $\times 4$) on U100.}
\label{tab:1}
\centering
\scalebox{0.90}[0.85]{
\begin{tabular}{|c|c|c|c|c|}
\hline
\multirow{2}{*}{Dataset} &
\multirow{2}{*}{Model} &
$\times 2$ & $\times 3$ & $\times 4$\\
\cline{3-5} & &PSNR/SSIM &PSNR/SSIM &PSNR/SSIM\\
\hline
\multirow{29}{*}{U100} &
Bicubic	&26.88/0.8403	&24.46/0.7349	&23.14/0.6577\\
\cline{2-5} &
A+\cite{timofte2014a+}	&29.20/0.8938	&26.03/0.7973	&24.32/0.7183\\
\cline{2-5} &
RFL\cite{schulter2015fast}	&29.11/0.8904	&25.86/0.7900	&24.19/0.7096\\
\cline{2-5} &
SelfEx\cite{huang2015single}	&29.54/0.8967	&26.44/0.8088 &24.79/0.7374\\
\cline{2-5} &
RED30\cite{mao2016image} &30.91/0.9159 &27.31/0.8303 &25.35/0.7587\\
\cline{2-5} &
DnCNN\cite{zhang2017beyond} &30.74/0.9139	&27.15/0.8276	&25.20/0.7521\\
\cline{2-5} &
TNRD\cite{chen2016trainable} &29.70/0.8994	&26.42/0.8076	&24.61/0.7291\\
\cline{2-5} &
FDSR\cite{lu2018fast} &30.91/0.9088	&27.23/0.8190	&25.27/0.7417\\
\cline{2-5} &
SRCNN\cite{dong2015image} &29.50/0.8946	&26.24/0.7989	&24.52/0.7221\\
\cline{2-5} &
FSRCNN\cite{dong2016accelerating} &29.88/0.9020	&26.43/0.8080	&24.62/0.7280\\
\cline{2-5} &
VDSR\cite{kim2016accurate} &30.76/0.9140	&27.14/0.8279	&25.18/0.7524\\
\cline{2-5} &
DRCN\cite{kim2016deeply} &30.75/0.9133	&27.15/0.8276	&25.14/0.7510\\
\cline{2-5} &
LapSRN\cite{lai2017deep} &30.41/0.9100	&-	&25.21/0.7560\\
\cline{2-5} &
IDN\cite{hui2018fast} &31.27/0.9196 &27.42/0.8359 &25.41/0.7632\\
\cline{2-5} &
DRRN\cite{tai2017image} &31.23/0.9188	&27.53/0.8378	&25.44/0.7638\\
\cline{2-5} &
BTSRN\cite{fan2017balanced} &\textcolor{blue}{31.63}/- &27.75/- &25.74- \\
\cline{2-5} &
MemNet\cite{tai2017memnet}	&31.31/0.9195	&27.56/0.8376	&25.50/0.7630\\
\cline{2-5} &
CARN-M\cite{ahn2018fast}	&30.83/\textcolor{blue}{0.9233}	&26.86/0.8263	
&25.63/0.7688\\
\cline{2-5} &
DRFN\cite{yang2018drfn} &31.08/0.9179 &27.43/0.8359 &25.45/0.7629\\
\cline{2-5} &
MADNet-${L_1}$\cite{lan2020madnet}  &31.62/0.9233	&27.77/\textcolor{blue}{0.8439}	&25.76/0.7746\\
\cline{2-5} &
MADNet-${L_F}$\cite{lan2020madnet} &31.59/\textcolor{red}{0.9234}	&\textcolor{blue}{27.78/0.8439}	&25.77/\textcolor{blue}{0.7751}\\
\cline{2-5} &
LESRCNN\cite{tian2020lightweight}	&31.45/0.9207	&27.76/0.8424	&\textcolor{blue}{25.78}/0.7739\\
\cline{2-5} &
DIP-FKP\cite{liang2021flow}  &24.33/0.7069  &23.47/0.6588 &22.89/0.6327\\
\cline{2-5} &
DIP-FKP+USRNet\cite{liang2021flow}  & 26.46/0.8203 & 24.84/0.7510 &23.89/0.7078\\
\cline{2-5} &
KOALAnet\cite{kim2021koalanet} &27.19/0.8318 &- &24.71/0.7427\\
\cline{2-5} &
FALSR-C\cite{chu2021fast} &   31.24/0.9187 &- &-\\
\cline{2-5} &
SRCondenseNet\cite{jain2018efficient} &31.24/0.9190 &- &-\\
\cline{2-5} &
SPSR\cite{ma2020structure} 
&24.80/0.9481 &- &-\\
\cline{2-5} &
DWSR\cite{guo2017deep}
&30.46/0.9162 &- &25.26/0.7548\\
\cline{2-5} &
S-BayeSR \cite{gao2022bayesian} &25.50/0.7528 &- &-\\
\cline{2-5} &
ESRGCNN (Ours) &\textcolor{red}{32.02}/0.9222 &\textcolor{red}{28.14/0.8512} &\textcolor{red}{26.10/0.7850}\\
\hline
\end{tabular}}
\label{tab:booktabs}
\end{table}
\subsection{Comparisons with state-of-the-arts}
To comprehensively test the performance of ESRGCNN, quantitative and qualitative analysis are chosen to conduct experiments in this paper. Specifically, the quantitative analysis uses SR results both of average PSNR and SSIM, running time of recovering high-quality image, model complexities and perceptual quality, i.e., feature similarity index (FSIM) \cite{zhang2011fsim} of popular SR methods, including Bicubic \cite{sun2008image}, A+ \cite{timofte2014a+}, RFL \cite{schulter2015fast}, self-exemplars super-resolution (SelfEx) \cite{huang2015single}, 30-layer residual encoder-decoder network (RED30) \cite{mao2016image}, the cascade of sparse coding based networks (CSCN) \cite{wang2015deep}, trainable nonlinear reaction diffusion (TNRD) \cite{chen2016trainable}, a denoising convolutional neural network (DnCNN) \cite{zhang2017beyond}, fast dilated super-resolution convolutional network (FDSR) \cite{lu2018fast}, SRCNN \cite{dong2015image}, residue context network (RCN) \cite{shi2017structure}, VDSR \cite{kim2016accurate}, context-wise network fusion (CNF) \cite{ren2017image}, Laplacian super-resolution network (LapSRN) \cite{lai2017deep}, information distillation network (IDN) \cite{hui2018fast}, DRRN \cite{tai2017image}, balanced two-stage residual networks (BTSRN) \cite{fan2017balanced}, MemNet \cite{tai2017memnet}, cascading residual network mobile (CARN-M) \cite{ahn2018fast}, end-to-end deep and shallow network (EEDS+) \cite{wang2019end}, deep recurrent fusion network (DRFN) \cite{yang2018drfn}, multiscale a dense lightweight network with L1 loss (MADNet-${L_1}$) \cite{lan2020madnet}, multiscale a dense lightweight network with enhanced LF loss (MADNet-${L_F}$) \cite{lan2020madnet}, multi-scale deep encoder-decoder with phase congruency (MSDEPC) \cite{liu2019single}, LESRCNN \cite{tian2020lightweight}, DIP-FKP\cite{liang2021flow},  DIP-FKP+USRNet\cite{liang2021flow}, kernel-oriented adaptive local adjustment network (KOALAnet)\cite{kim2021koalanet},
fast, accurate and lightweight super-resolution architectures and models (FALSR-C)\cite{chu2021fast}, SRCondenseNet\cite{jain2018efficient}, structure-preserving super resolution method (SPSR)\cite{ma2020structure}, 
residual dense network (RDN) \cite{zhang2018residual}, channel-wise and spatial feature modulation (CSFM) \cite{hu2019channel}, super-resolution feedback network (SRFBN) \cite{li2019feedback}, deep wavelet super-resolution
(DWSR)\cite{guo2017deep}, S-BayeSR \cite{gao2022bayesian},  coarse-to-fine super-resolution CNN (CFSRCNN) \cite{tian2020coarse} on four benchmark datasets, i.e., Set5 \cite{bevilacqua2012low}, Set14 \cite{bevilacqua2012low}, B100 \cite{martin2001database} and U100 \cite{huang2015single} to verify the SISR performance of the ESRGCNN. Quantitative analysis uses predicted SR images of different methods to visually test the SR effect of ESRGCNN.

Quantitative analysis: We select  indexes of PSNR and SSIM
on Set5, Set14, B100 and U100 to verify the SR performance of different techniques in Tables 4-7, where red and blue lines
are symbolized as the highest and second SISR results, respectively.

In Table 4, we see that ESRGCNN is superior to other popular SR methods, i.e., CARN-M, MADNet-${L_F}$ and LESRCNN in terms of PSNR on Set5 for $\times 3$ and  $\times 4$ upscaling, respectively. Also, as illustrated in Tables 4-7, ESRGCNN almost achieves the best performance for all the three scales (i.e., $\times 2$,  $\times 3$ and $\times 4$). For instance, the ESRGCNN obtains a notable gain of 0.12dB in PSNR and 0.0020 in SSIM on Set14 than that of the second MADNet-${L_F}$ for $\times 4$  upscaling in Table 5. Also, ESRGCNN implements an excellent PSNR gain over the second MADNet-${L_F}$ by 0.36dB and a significant SSIM gain over the second MADNet-${L_F}$ by 0.0073 on U100 for $\times 3$  upscaling as reported in Table 7. Besides, the proposed ESRGCNN outperforms the state-of-the-arts, including RDN, CSFM, SRFM and CFSRCNN on large dataset of B100 for SISR as given in Table 8, where the best and second SISR performance are remarked as red line and blue line, respectively. To verify validity of our ESRGCNN, we choose  deep network cascade (DNC) \cite{li2019feedback},
fuzzy deep CNN (FDCNN)\cite{greeshma2017single}, adaptive network based fuzzy inference system (ANFIS) \cite{ismail2020image} and image super-resolution method via a weighted random forest model (SWRF) \cite{liu2017image} as comparative methods for an image super-resolution with a certain scene. As shown in Tables 9 and 10, our proposed method is superior to other popular super-resolution methods in certain scene. These materials above prove that the proposed ESRGCNN has good robustness for recovering LR images of different background.
\begin{figure}[!t]
\centering
\subfloat{\includegraphics[width=3.5in]{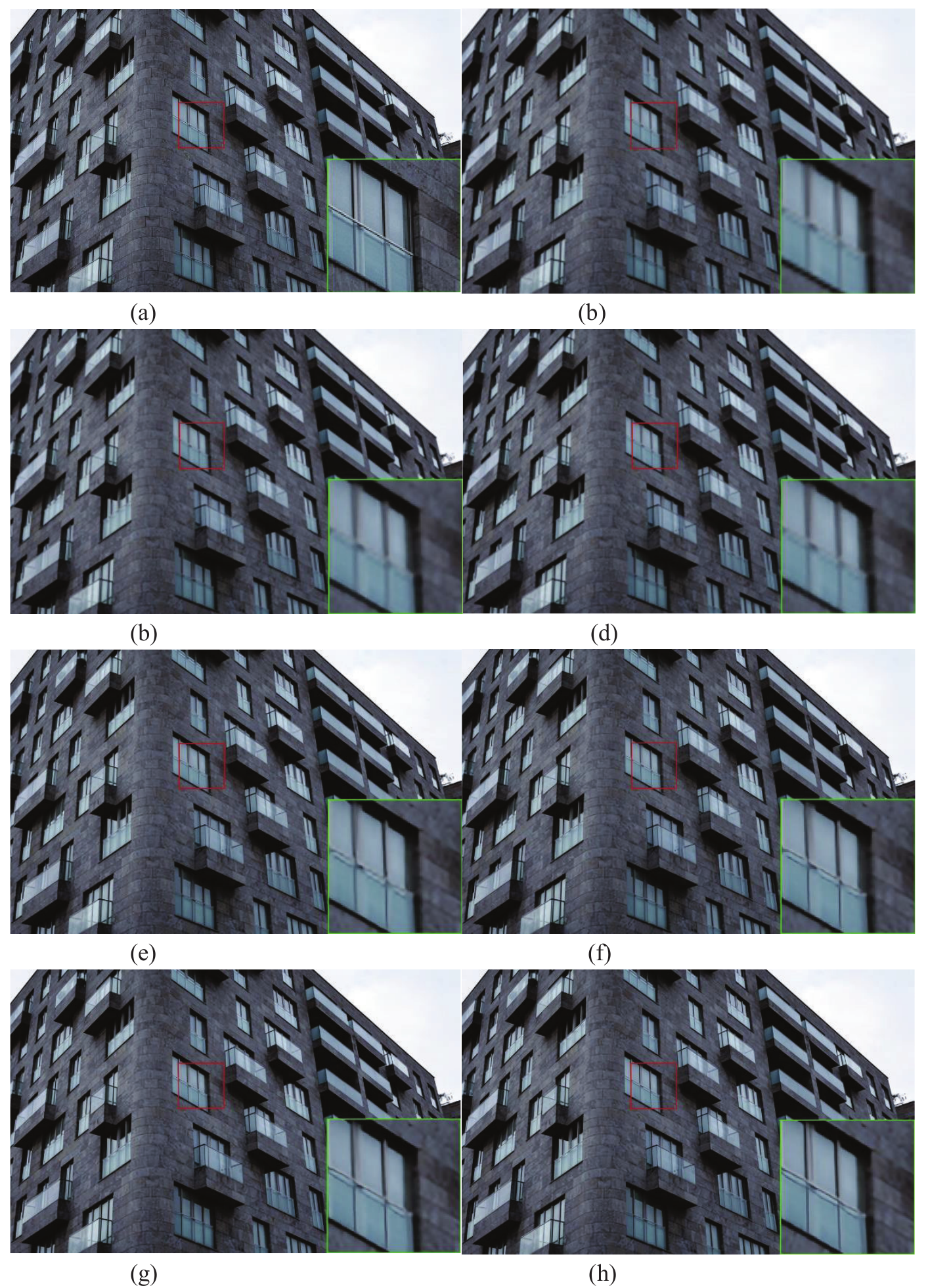}
\label{fig_second_case}}
\caption{Visual effect of different methods for ×3 upscaling on U100. Also, obtained PSNR and SSIM of these methods are given as follows. (a) Bicubic (26.19 dB/0.7295), (b) VDSR (28.44 dB/0.8077), (c) DRCN (28.40 dB/0.8074), (d) CARN-M (28.90 dB/0.8171), (e) LESRCNN (29.06 dB/0.8199), (f) CFSRCNN (29.55 dB/0.8298), (g) ACNet (29.53 dB/0.8289) and (h) ESRGCNN (Ours) (29.58 dB/0.8303).}
\label{fig:7}
\end{figure}

It is known that execution time and complexity are important metrics for digital devices \cite{tian2020coarse}. Inspired by that, using six popular SR methods, i.e., VDSR, DRRN, MemNet, RDN, SRFBN and CRAN-M to predict high-quality images of different sizes (i.e., $256 \times 256$, $512 \times 512$ and $1024 \times 1024$) with $\times 2$ upscaling for testing running time of these methods. From Table 11, we can see that the ESRGCNN obtains the fastest recovering speed in SISR. Also, eight SR networks (i.e., VDSR, DnCNN, DRCN, MemNet, CSFM, RDN, SRFBN and CFSRCNN) are used to conduct experiments for comparing complexity of ESRGCNN. Specifically, total parameters and flops \cite{tian2020deep} in Table 12 are denoted as complexity of computational cost and memory consumption for predicting SR images of size $166 \times 166$. Table 10 reports the ESRGCNN takes the fewest number of flops for restoring HR images. Besides, FSIM value of perceptual quality is used to evaluate the visual effect of different SR. Table 13 shows that our ESRGCNN achieves better effects in FSIM values in comparison with other SR methods on B100 for $\times 2$, $\times 3$ and $\times 4$, respectively. Besides, the best and second performance are listed by red line and blue line in Tables 11-13, respectively. 

\begin{table}[t!]
\caption{Average PSNR/SSIM results of different SR methods for $\times 4$ upscaling on B100.}
\label{tab:1}
\centering
\scalebox{0.9}[0.9]{
\begin{tabular}{|c|c|c|}
\hline
Methods	&PSNR(dB)	&SSIM\\
\hline
RDN	\cite{zhang2018residual} &27.72	&0.7419\\
\hline
CSFM \cite{hu2019channel}	&27.76	&0.7432\\
\hline
SRFBN \cite{li2019feedback}	&27.72	&0.7409\\
\hline
CFSRCNN	\cite{tian2020coarse} &\textcolor{blue}{27.53}	&\textcolor{blue}{0.7333}\\
\hline
ESRGCNN (Ours) &\textcolor{red}{27.57} &\textcolor{red}{0.7348}\\
\hline
\end{tabular}}
\label{tab:booktabs}
\end{table}

\begin{table}[t!]
\caption{Performance of different SR methods on a parrot image for $\times 3$ upscaling.}
\label{tab:1}
\centering
\scalebox{0.8}[0.8]{
\begin{tabular}{|c|c|c|}
\hline
Methods &PSNR &SSIM\\
\hline
DNC \cite{cui2014deep} &30.18 &0.913\\
\hline
FDCNN \cite{greeshma2017single} &35.47
&0.957\\
\hline
ESRGCNN (Ours) &37.69 &0.969\\
\hline
\end{tabular}}
\label{tab:booktabs}
\end{table}

\begin{table}[t!]
\caption{Performance of different SR methods on a butterfly image for $\times 2$ upscaling.}
\label{tab:1}
\centering
\scalebox{0.8}[0.8]{
\begin{tabular}{|c|c|c|}
\hline
Methods &PSNR &SSIM \\
\hline
ANFIS \cite{ismail2020image} &29.94 &0.942\\
\hline
SWRF \cite{liu2017image} &32.19
&-\\
\hline
ESRGCNN (Ours) &39.80 &0.981\\
\hline
\end{tabular}}
\label{tab:booktabs}
\end{table}

\begin{figure*}[!t]
\centering
\subfloat{\includegraphics[width=4.0in]{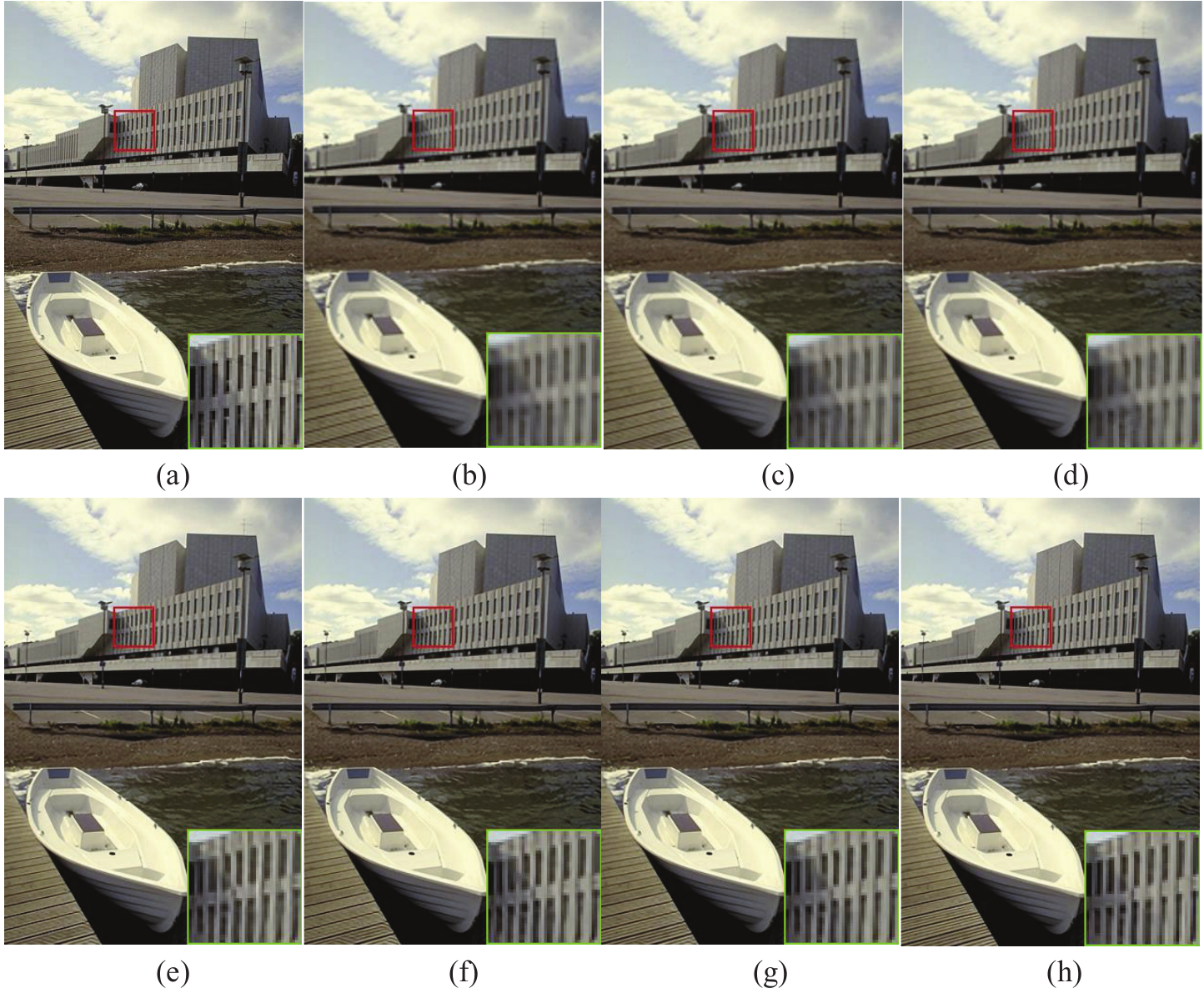}
\label{fig_second_case}}
\caption{Visual effect of different methods for ×2 upscaling on B100. Also, obtained PSNR and SSIM of these methods are given as follows. (a) Bicubic (27.63 dB/0.8220), (b) VDSR (30.50 dB/0.8956), (c) DRCN (30.47 dB/0.8938), (d) CARN-M (30.60 dB/0.8967), (e) LESRCNN (30.94 dB/0.8987), (f) CFSRCNN (31.43 dB/0.9067), (g) ACNet (31.24 dB/0.9036) and (h) ESRGCNN (Ours) (31.56 dB/0.9085).}
\label{fig:7}
\end{figure*}

\begin{table}[t!]
\caption{Running time (s) of different SR methods on recovering HR images of sizes $256\times256$, $512\times512$ and $1024\times1024$ for $\times2$ upcaling.}
\label{tab:1}
\centering
\scalebox{0.74}[0.76]{
\begin{tabular}{|c|c|c|c|}
\hline
\multicolumn{4}{|c|}{Single image super-resolution} \\
\hline
Size &$256 \times 256$ &$512 \times 512$ &$1024 \times 1024$\\
\hline
VDSR\cite{kim2016accurate} &0.0172	&0.0575  &0.2126\\
\hline
DRRN\cite{tai2017image}  &3.063 	&8.050  &25.23\\
\hline
MemNet\cite{tai2017memnet} &0.8774		&3.605   &14.69\\
\hline
RDN \cite{zhang2018residual} &0.0553 &0.2232 &0.9124\\
\hline
SRFBN \cite{li2019feedback} &0.0761 &0.2508 &0.9787\\
\hline
CARN-M\cite{ahn2018fast} &\textcolor{blue}{0.0159} &\textcolor{blue}{0.0199} &\textcolor{blue}{0.0320}\\
\hline
ESRGCNN (Ours) &\textcolor{red}{0.0157} &\textcolor{red}{0.0185} &\textcolor{red}{0.0294}\\
\hline
\end{tabular}}
\label{tab:booktabs}
\end{table}

\begin{table}[t!]
\caption{Complexities of different SR methods for $\times 2$ upscaling.}
\label{tab:1}
\centering
\scalebox{0.8}[0.8]{
\begin{tabular}{|c|c|c|}
\hline
Methods	&Parameters	&Flops\\
\hline
VDSR\cite{kim2016accurate}	&\textcolor{blue}{665K}	&18.32G\\
\hline
DnCNN\cite{zhang2017beyond}	&\textcolor{red}{556K}	&15.32G\\
\hline
DRCN\cite{kim2016deeply}	&1,774K	&48.88G\\
\hline
MemNet\cite{tai2017memnet}	&677K	&18.66G\\
\hline
CSFM \cite{hu2019channel} &12,841K &89.26G\\
\hline
RDN \cite{zhang2018residual} &21,937K &151.92G\\
\hline
SRFBN \cite{li2019feedback} &3,631K &26.57G\\
\hline
CFSRCNN \cite{tian2020coarse} &1,200K &\textcolor{blue}{13.64G}\\
\hline
ESRGCNN (Ours) &1,238K &\textcolor{red}{9.33G}\\
\hline
\end{tabular}}
\label{tab:booktabs}
\end{table}

\begin{table}[htbp!]
\caption{FSIM values of different SR methods for $\times 2$, $\times 3$ and $\times 4$ upscaling on B100.}
\label{tab:1}
\centering
\scalebox{0.79}[0.80]{
\begin{tabular}{|c|c|c|c|c|}
\hline
\multirow{1}{*}{Dataset} &
\multirow{1}{*}{Methods} &
$\times 2$ & $\times 3$ & $\times 4$\\
\hline
\multirow{6}{*}{B100} &
A+\cite{timofte2014a+}	&0.9851	&0.9734	&0.9592\\
\cline{2-5} &
SelfEx\cite{huang2015single} &0.9976	&0.9894	&0.9760\\
\cline{2-5} &
SRCNN\cite{dong2015image} &0.9974	&0.9882	&0.9712\\
\cline{2-5} &
CARN-M\cite{ahn2018fast} &\textcolor{blue}{0.9979}	&0.9898	&0.9765\\
\cline{2-5} &
LESRCNN\cite{tian2020lightweight} &\textcolor{blue}{0.9979}	&\textcolor{blue}{0.9903}	&\textcolor{blue}{0.9774}\\
\cline{2-5} &
ESRGCNN (Ours) &\textcolor{red}{0.9980} &\textcolor{red}{0.9905} &\textcolor{red}{0.9777}\\
\hline
\end{tabular}}
\label{tab:booktabs}
\end{table}

Qualitative analysis: Four SR methods (i.e., Bicubic, CARN-M, LESRCNN and ESRGCNN) on U100 for $\times 3$ upscaling and B100 for $\times 2$ upscaling are used to construct high-quality images, respectively. To easier observe definition of predicted SR images of different methods, one area of the predicted SR image is amplified as observation area. It is known that the observation area has higher clarity, the corresponding SR method has better SR effect. Figs. 4-5 show that selected regions by ESRGCNN are clearer than that of other SR methods, which show the ESRGCNN is very competitive in visual effect of SISR. According to mentioned quantitative analysis and qualitative analysis, we can see that the proposed ESRGCNN is very suitable to SISR on digital devices.

\section{Conclusion}
This paper presents an enhanced super-resolution group CNN (ESRGCNN) for SISR. ESRGCNN enhances the effect of deep and wide channel features by correlations of different channels to extract more accurate low-frequency information for SISR. Also, taking long-term dependence problem of deep network into consideration, a signal enhancement operation is fused into ESRGCNN for inheriting more long-distance contextual information. Besides, to deal with low-resolution images of different sizes, an adaptive up-sampling operation is applied to achieve a SR model. Comprehensive experiments on several benchmark datasets prove that ESRGCNN achieves an excellent effect among SISR results, SISR efficiency, SR model complexity and visual quality. We will use signal processing techniques, math ideas and deep learning theory to design lightweight CNNs for blind image super-resolution in the future. 

\section*{Acknowledgments}
This work was supported  in part by the Fundamental Research Funds for the Central Universities, China under Grant D5000210966, Shenzhen-Hong Kong Innovation Circle
Category D Project, China SGDX2019081623300177 (CityU 9240008), and in part by Ministry of science and
Technology, Taiwan, under Grant 110-2634-F-007-015-.


\bibliographystyle{elsarticle-harv}
\bibliography{references}

\end{spacing}
\end{document}